%% file: SCMaree_HillVallEA_Niching_Competition_2019.tex
\documentclass[10pt,a4paper,twoside]{article}
\usepackage{booktabs}

\usepackage[utf8]{inputenc} 
\usepackage{amsmath} 
\usepackage{amsfonts} 
\usepackage{amssymb} 
\usepackage{amsthm} 
\usepackage{graphicx} 
\usepackage{epstopdf} 
\usepackage{bbm} 
\usepackage[lined,boxed,commentsnumbered]{algorithm2e}
\usepackage[toc]{appendix} 
\usepackage[headsep=1cm,headheight=0cm]{geometry}
\usepackage{enumerate}
\usepackage{listings}
\usepackage{authblk}

	\usepackage[pdfborder={0 0 0},plainpages=false]{hyperref}

	\theoremstyle{plain}

	\theoremstyle{remark}

\title{Benchmarking HillVallEA for the GECCO 2019 Competition on Multimodal Optimization}

\input{math_shortcuts}

\date{\today}
\author[1]{S.C. Maree}
\author[1]{T. Alderliesten}
\author[2]{P.A.N. Bosman}

\affil[1]{Amsterdam UMC, Amsterdam, The Netherlands}
\affil[2]{Centrum Wiskunde \& Informatica, Amsterdam, The Netherlands}

\begin{document}

\maketitle

\begin{abstract}
This report presents benchmarking results of the Hill-Valley Evolutionary Algorithm version 2019 (HillVallEA19) on the CEC2013 niching benchmark suite under the restrictions of the GECCO 2019 niching competition on multimodal optimization. Performance is compared to algorithms that participated in previous editions of the niching competition.
\end{abstract}

\section{Introduction}
The Hill-Valley Evolutionary Algorithm (HillVallEA) \cite{Maree18, Maree18b} is a real-valued multi-modal evolutionary algorithm, that automatically detects niches in the search space, based on the Hill-Valley test. This test states that two solutions belong to the same niche (valley) when there is no hill in between. To do so, a number of intermediate solutions are sampled and evaluated. Hill-Valley Clustering (HVC) is an iterative approach to efficiently cluster an entire population of solutions into niches. The resulting clusters are used to initialze a population-based core search algorithm, in this case, AMaLGaM-Univariate \cite{bosman08} is used.

\section{Adaptations to the HillVallEA19}
Some small adaptations have been made to HillVallEA18 \cite{Maree18b} to further enhance its performance. Source code of HillVallEA is available at \url{github.com/scmaree/HillVallEA}.

\subsection{Adaptive initial population sampling}
In HillVallEA, after all local optimizers have been terminated (and there is still budget remaining), a new initial population is sampled. When this population is sampled uniformly random, previously explored basins get re-explored every time a new population is initialized. To reduce this computational overhead, we store for each solution of the previous initial population to which cluster it belonged.  Then, new solutions are sampled based on rejection sampling. A sample is rejected with probability $P = 0.9$ if its nearest $d+1$ solutions of the previous initial population belonged to the same cluster. In that case, it is very likely that this solution will end up exploring the same basin as the cluster of the previous generation. 

Additionally, better spreading the initially sampled population has been shown to improve performance of evolutionary algorithms \cite{wessing15}. Minimax sampling method or Latin hypercube sampling \cite{wessing15} grow slow very quickly as the problem dimensionality or the sample size increase. We therefore use a greedy scattered subset selection method \cite{rodrigues14}. To construct a population of $N$ solutions, we sample $2N$ using the strategy above, and use greedy scattered subset selection to reduce this to $N$ solutions.

\subsection{Force accept of low-fitness solutions in Hill-Valley test}
Previously in hill-valley clustering, all nearest-better solutions were tested with the hill-valley test. Especially for problems with a large number of low-fitness local optima, such as the Shubert function (problems 6 and 8 in the benchmark), this results in many resources spent on obtaining accurate low-fitness clusters that are later discarded. Therefore, during hill-valley clustering, the hill-valley test is only performed on solutions that belong to the fittest half of the selection, or on solutions-pairs that are more than the expected edge length (EEL) \cite{Maree18} apart (i.e., $N_t > 1$).

\subsection{Recalibration of the recursion scheme}
The population sizing scheme within HillVallEA is parameterized as $\xi = (N,N^{\mbox{inc}}, N_C,N_C^{\mbox{inc}}, )$,  with initial population size $N$, population size increment $N^{\mbox{inc}}$, cluster size $N_C$, and cluster size increment $N_C^{\mbox{inc}}$. Previously, these parameters were set to $\xi = (2^8d, 2, 1, 1.2)$, where $d$ is the problem dimensionality. As the initial population sampling has been adapted, setting $\xi = (2^6, 2, 0.8, 1.1)$ was found to enhance performance of the HillVallEA. That is, using both a smaller population and smaller clusters initially, and increasing cluster size at a slower pace.

\section{Experiment Setup}
We evaluate the performance of HillVallEA on the test problems in the CEC2013 niching benchmark suite \cite{CEC2013NichingCompetition}. The benchmark consists of 20 problems, as shown in Table~\ref{tab:2dbenchmarks}, to be solved within a predefined budget in terms of function evaluations. For each of the benchmark problems, the location of the optima and the corresponding fitness values are known, however, these are only used for to measure performance, and not used during optimization. All benchmark functions are defined on a bounded domain. All experiments are repeated 50 times, and resulting performance measures are averaged over all repetitions. Note that no problem-specific parameter tuning has been performed.

\subsection{Performance Metrics}
\label{sec:measures}
Two performance measures are used, from which three scoring scenarios are computed, according to the competition guidelines. Let $\cO$ be the set of presumed optima obtained by an algorithm, and let $g$ be the number of distinct global optima within $\cO$. Finally, let $G_p$ be the number of global optima for problem $p$. Then, we define the peak ratio (PR) as $\mbox{PR} = g / G_p$ and the success rate (SR) as $\mbox{SR} = g / |\cO|$. Both measures should be maximized, with maximum $1$. From these two measures, three scoring scenarios are constructed. 

\begin{enumerate}
\item[S1] The first scenario is simply the PR. 
\item[S2] The second scenario is known as the static $F_1$ measure, defined as $F_1 = \frac{2 \cdot PR \cdot SR}{PR + SR}$.
\item[S3] The third and final scenario is the dynamic $F_1$ (dyn$F_1$), which is the area under the curve of the $F_1$ over time (in number of function evaluations). For this, sort the solutions in $\cO$ based on the number of function evaluations $f_i$ before a solution $o_i\in\cO$ was considered a global optimum, with the first-obtained solution first. Let $\cO_{[1:t]}$ with $t\in[1,|\cO|]$ be the subset of $\cO$ containing the first $t$ solutions and let $B_p$ be the function evaluation budget for problem $p$. Then we can write the dyn$F_1$ as,
$$\mbox{dyn}F_1 =  \left(\frac{B_p - f_{|\cO|}}{B_p}\right) F_1(\cO)  + \sum_{i = 2}^{|\cO|} \left( \frac{f_i - f_{i-1} }{B_p}\right) F_1(\cO_{[1:i-1]}).$$
\end{enumerate}

According to the competition guidelines, a solution is marked as a distinct global optima for five different accuracy levels $\varepsilon = \{10^{-1}, 10^{-2}, 10^{-3}, 10^{-4}, 10^{-5}\}$. Then, for each of the accuracy levels, for each problem, the scenario score is the average of the scores over the five accuracy levels.

\subsection{Algorithms}
\label{sec:algorithms}
We compare performance of HillVallEA-2019 to all algorithms that previously participated in the niching competitions held at the GECCO and CEC conferences in 2016, 2017 and 2018. The raw solution sets are used. The obtained solutions are re-evaluated and the scores under the different scenarios are computed given the definitions stated above. Note that the algorithms are not re-run. The included algorithms are, 
NEA2+ \cite{preuss12}, RLSIS \cite{wessing15}, RS-CMSA \cite{ahrari17} , HillVallEA18 \cite{Maree18b}, SDE-Ga (No known reference, developed by Jun-ichi Kushida) and finally the method that we discus in this paper, HillVallEA19.

\section{Results and discussion}
Tables~\ref{tab:s1}, \ref{tab:s2} and \ref{tab:s3} shows the score per problem and per algorithm under respectively scenario S1, S2 and S3. Table~\ref{tab:ranks} shows the overall score and the corresponding ranks. 

The SR of HillVallEA is 1 in all cases, and almost always for RS-CMSA, which shows that the similar post-processing step that both algorithms perform is successful at removing duplicates and local optima. Problems 1-5 and 10 are fully solved by all methods in all runs for all accuracy levels (except for two runs of NEA2+), which suggesting that these problems are too simple. 

No method can obtain the final two Weierstrass peaks of Composition Function 4 fully in any dimension (problem 15, 17, 19 and 20), which may indicate that this is a needle in a haystack problem. Similarly to problem 18, for which the obtained peak ratio was 0.667 for all methods (except NEA2+). Especially, SDE-Ga obtains for problems 13, 14, 16, and 18 a maximum peak ratio of 0.667, being unable to solve the Weierstrass function. 

To conclude, HillVallEA19 was shown to be an improvement over HillVallEA18 under all scenarios. RS-CMSA comes directly after in all three scenarios. SDE-Ga performs well under S1 and S2, especially for problems 8 and 9, but performance deteriorates for the higher-dimensional problems.  SDE-Ga obtains solutions very late in the convergence process, resulting in a very low S3 score. Overall, HillVallEA19 performs best under all scenarios.

\begin{table}
\begin{center}
\caption{Niching benchmark suite from the CEC2013 special session on multi-modal optimization \cite{CEC2013NichingCompetition}. For each problem the function name, problem dimensionality $d$, number of global optima $\#gopt$, and local optima $\#lopt$ and budget in terms of function evaluations are given.}
\label{tab:2dbenchmarks}
\small
\begin{tabular}{cccccc}
\toprule
\# & Function name & $d$ & \#gopt & \#lopt & budget \\
\toprule
1 & Five-Uneven-Peak Trap & 1 & 2 & 3 & 50K  \\
2 & Equal Maxima & 1 & 5 & 0 & 50K  \\
3 & Uneven Decreasing Maxima & 1 & 1 & 4 & 50K \\
4 & Himmelblau & 2 & 4 & 0 & 50K\\
5 & Six-Hump Camel Back & 2 & 2 & 5 & 50K \\
6 & Shubert & 2 & 18 & many & 200K   \\
7 & Vincent & 2 & 36 & 0 & 200K \\
8 & Shubert & 3 & 81 & many & 400K  \\
9 & Vincent & 3 & 216 & 0 & 400K  \\
10 & Modified Rastrigin & 2 & 12 & 0 & 200K  \\
11 & Composition Function 1 & 2 & 6 & many &200K \\
12 & Composition Function 2 & 2 & 8 &many & 200K \\
13 & Composition Function 3 & 2 & 6 &many & 200K  \\
14 & Composition Function 3 & 3 & 6 &many & 400K  \\
15 & Composition Function 4 & 3 & 8 &many & 400K  \\
16 & Composition Function 3 & 5 & 6 &many & 400K \\
17 & Composition Function 4 & 5 & 8 &many & 400K \\
18 & Composition Function 3 & 10 & 6 & many &400K  \\
19 & Composition Function 4 & 10 & 8 &many & 400K  \\
20 & Composition Function 4 & 20 & 8 &many & 400K \\
\bottomrule
\end{tabular}
\end{center}
\end{table}

\begin{table}
\begin{center}
\caption{Scores obtained under Scenario S1 (peak ratio) for each of the algorithms per problem $p$. Higher is better, 1 is the maximum score. Scores are averaged over 50 runs and five accuracy levels. Average (avg.) score computed over all 20 problems. }
\label{tab:s1}
\smaller
\begin{tabular}{c|cccc|cc}
\toprule
& & &  & & \multicolumn{2}{|c}{HillVallEA} \\
  p & NEA2+ & RLSIS & RS-CMSA & SDE-Ga & HillVallEA18 & HillVallEA19 \\ 
\toprule
   1 & 1.000 & 1.000 & 1.000 & 1.000 & 1.000 & 1.000 \\
   2 & 1.000 & 1.000 & 1.000 & 1.000 & 1.000 & 1.000 \\
   3 & 1.000 & 1.000 & 1.000 & 1.000 & 1.000 & 1.000 \\
   4 & 0.998 & 1.000 & 1.000 & 1.000 & 1.000 & 1.000 \\
   5 & 1.000 & 1.000 & 1.000 & 1.000 & 1.000 & 1.000 \\
   6 & 0.997 & 0.872 & 0.999 & 1.000 & 1.000 & 1.000 \\
   7 & 0.840 & 0.920 & 0.997 & 1.000 & 1.000 & 1.000 \\
   8 & 0.568 & 0.189 & 0.871 & 1.000 & 0.920 & 0.975 \\
   9 & 0.552 & 0.584 & 0.730 & 0.992 & 0.945 & 0.972 \\
  10 & 0.997 & 1.000 & 1.000 & 1.000 & 1.000 & 1.000 \\
  11 & 0.955 & 1.000 & 0.997 & 0.733 & 1.000 & 1.000 \\
  12 & 0.796 & 0.950 & 0.948 & 0.800 & 1.000 & 1.000 \\
  13 & 0.947 & 0.938 & 0.997 & 0.667 & 1.000 & 1.000 \\
  14 & 0.813 & 0.799 & 0.810 & 0.667 & 0.917 & 0.923 \\
  15 & 0.721 & 0.720 & 0.748 & 0.750 & 0.750 & 0.750 \\
  16 & 0.683 & 0.675 & 0.667 & 0.667 & 0.687 & 0.723 \\
  17 & 0.723 & 0.738 & 0.703 & 0.703 & 0.750 & 0.750 \\
  18 & 0.650 & 0.667 & 0.667 & 0.667 & 0.667 & 0.667 \\
  19 & 0.505 & 0.515 & 0.502 & 0.555 & 0.585 & 0.593 \\
  20 & 0.398 & 0.422 & 0.482 & 0.460 & 0.482 & 0.480 \\

\bottomrule
 avg  & 0.807 & 0.800 & 0.856 & 0.833 & 0.885 & 0.892 \\
\bottomrule
\end{tabular}
\end{center}
\end{table}

\begin{table}
\begin{center}
\caption{Scores obtained under Scenario S2 (static $F_1$) for each of the algorithms per problem $p$. Higher is better, 1 is the maximum score. Scores are averaged over 50 runs and five accuracy levels. Average (avg.) score computed over all 20 problems. }
\label{tab:s2}
\smaller
\begin{tabular}{c|cccc|cc}
\toprule
& & &  & & \multicolumn{2}{|c}{HillVallEA} \\
  \# & NEA2+ & RLSIS & RS-CMSA & SDE-Ga & HillVallEA18 & HillVallEA19 \\ 
\toprule
   1 & 1.000 & 0.993 & 0.996 & 1.000 & 1.000 & 1.000 \\
   2 & 1.000 & 0.993 & 1.000 & 1.000 & 1.000 & 1.000 \\
   3 & 1.000 & 0.993 & 0.987 & 1.000 & 1.000 & 1.000 \\
   4 & 0.960 & 0.978 & 1.000 & 1.000 & 1.000 & 1.000 \\
   5 & 0.947 & 0.949 & 1.000 & 1.000 & 1.000 & 1.000 \\
   6 & 0.997 & 0.924 & 0.999 & 1.000 & 1.000 & 1.000 \\
   7 & 0.614 & 0.947 & 0.999 & 1.000 & 1.000 & 1.000 \\
   8 & 0.723 & 0.315 & 0.931 & 1.000 & 0.958 & 0.987 \\
   9 & 0.646 & 0.733 & 0.844 & 0.996 & 0.972 & 0.986 \\
  10 & 0.997 & 0.988 & 1.000 & 1.000 & 1.000 & 1.000 \\
  11 & 0.971 & 0.992 & 0.998 & 0.733 & 1.000 & 1.000 \\
  12 & 0.881 & 0.967 & 0.972 & 0.800 & 1.000 & 1.000 \\
  13 & 0.966 & 0.941 & 0.998 & 0.723 & 1.000 & 1.000 \\
  14 & 0.894 & 0.865 & 0.893 & 0.799 & 0.953 & 0.958 \\
  15 & 0.835 & 0.831 & 0.855 & 0.857 & 0.857 & 0.857 \\
  16 & 0.811 & 0.795 & 0.800 & 0.800 & 0.813 & 0.837 \\
  17 & 0.838 & 0.843 & 0.823 & 0.824 & 0.857 & 0.857 \\
  18 & 0.787 & 0.794 & 0.800 & 0.800 & 0.800 & 0.800 \\
  19 & 0.668 & 0.676 & 0.668 & 0.712 & 0.735 & 0.741 \\
  20 & 0.563 & 0.590 & 0.650 & 0.627 & 0.650 & 0.647 \\
\bottomrule
 avg  & 0.855 & 0.855 & 0.911 & 0.884 & 0.930 & 0.934 \\
\bottomrule
\end{tabular}
\end{center}
\end{table}

\begin{table}
\begin{center}
\caption{Scores obtained under Scenario S3 (dynamic $F_1$) for each of the algorithms per problem $p$. Higher is better, 1 is the maximum score. Scores are averaged over 50 runs and five accuracy levels. Average (avg.) score computed over all 20 problems. }
\label{tab:s3}
\smaller
\begin{tabular}{c|cccc|cc}
\toprule
& & &  & & \multicolumn{2}{|c}{HillVallEA} \\
  \# & NEA2+ & RLSIS & RS-CMSA & SDE-Ga & HillVallEA18 & HillVallEA19 \\ 
\toprule
   1 & 0.982 & 0.987 & 0.911 & 0.572 & 0.992 & 0.995 \\
   2 & 0.917 & 0.983 & 0.959 & 0.626 & 0.987 & 0.989 \\
   3 & 0.895 & 0.982 & 0.949 & 0.456 & 0.992 & 0.994 \\
   4 & 0.959 & 0.975 & 0.932 & 0.528 & 0.973 & 0.977 \\
   5 & 0.971 & 0.979 & 0.944 & 0.416 & 0.982 & 0.983 \\
   6 & 0.917 & 0.699 & 0.933 & 0.525 & 0.951 & 0.966 \\
   7 & 0.659 & 0.855 & 0.928 & 0.413 & 0.960 & 0.966 \\
   8 & 0.464 & 0.209 & 0.715 & 0.324 & 0.750 & 0.805 \\
   9 & 0.550 & 0.573 & 0.654 & 0.157 & 0.791 & 0.818 \\
  10 & 0.988 & 0.983 & 0.984 & 0.539 & 0.979 & 0.982 \\
  11 & 0.961 & 0.980 & 0.967 & 0.254 & 0.983 & 0.983 \\
  12 & 0.829 & 0.859 & 0.909 & 0.303 & 0.958 & 0.963 \\
  13 & 0.932 & 0.897 & 0.923 & 0.372 & 0.957 & 0.964 \\
  14 & 0.862 & 0.834 & 0.832 & 0.434 & 0.867 & 0.882 \\
  15 & 0.806 & 0.782 & 0.785 & 0.544 & 0.824 & 0.836 \\
  16 & 0.799 & 0.788 & 0.777 & 0.449 & 0.776 & 0.793 \\
  17 & 0.803 & 0.779 & 0.688 & 0.419 & 0.787 & 0.816 \\
  18 & 0.719 & 0.748 & 0.730 & 0.160 & 0.721 & 0.763 \\
  19 & 0.624 & 0.627 & 0.560 & 0.035 & 0.634 & 0.656 \\
  20 & 0.491 & 0.496 & 0.502 & 0.003 & 0.514 & 0.524 \\
\bottomrule
 avg  & 0.806 & 0.801 & 0.829 & 0.376 & 0.869 & 0.883 \\
\bottomrule
\end{tabular}
\end{center}
\end{table}

\begin{table}
\begin{center}
\caption{Algorithm ranks based on the three scenarios.  }
\label{tab:ranks}
\smaller
\begin{tabular}{c|cc|cc|cc|c}
\toprule
Algorithm & S1 & rank & S2 & rank & S3 & rank & average rank \\
\toprule
                 RLSIS & 0.800 & 6& 0.855 &5  & 0.801 & 5& 5.3\\
                 NEA2+ & 0.807 & 5& 0.855 & 6  & 0.806 &4 & 5\\ 
                SDE-Ga & 0.833 & 4& 0.884 & 4 & 0.376 & 6& 4.7\\  
           RS-CMSA & 0.856 &3 & 0.911 &  3& 0.829 & 3& 3\\ 
          HillVallEA18 & 0.885 &2 & 0.930 & 2 & 0.869 &2 & 2\\ 
          HillVallEA19 & 0.892& 1& 0.934 &  1 & 0.883 & 1& 1\\
\bottomrule
\end{tabular}
\end{center}
\end{table}

\bibliographystyle{acm}
\bibliography{Maree_2019}

\end{document}

%% file: math_shortcuts.tex
\newcommand{\cO}{\mathcal{O}}

\makeatletter
\def\Ddots{\mathinner{\mkern1mu\raise\p@
\vbox{\kern7\p@\hbox{.}}\mkern2mu
\raise4\p@\hbox{.}\mkern2mu\raise7\p@\hbox{.}\mkern1mu}}
\makeatother